\documentclass[runningheads]{llncs}
\usepackage{graphicx}
\usepackage{amsmath,amssymb} % define this before the line numbering.
\usepackage{color}
\usepackage[width=122mm,left=12mm,paperwidth=146mm,height=193mm,top=12mm,paperheight=217mm]{geometry}
\usepackage[rmdefault=mdbch,sfdefault=zavm,OMLmathrm,OMLmathbf,OMLmathsfit]{isomath}
\usepackage{siunitx}
\usepackage{xifthen}
\usepackage[mathcal]{eucal}
\usepackage{scalerel}
\usepackage{mathtools}
\usepackage{url}

\renewcommand{\vec}[1]{\mathbfit{#1}}

\newcommand{\mat}[1]{\mathbfit{#1}}

\newcommand{\randmat}[1]{\mathbf{#1}}

\begin{document}
\pagestyle{headings}
\mainmatter

\title{Siamese Generative Adversarial Privatizer\\for Biometric Data} 

\titlerunning{Siamese Generative Adversarial Privatizer\\for Biometric Data}

\authorrunning{W. Oleszkiewicz, P. Kairouz, K. Piczak, R. Rajagopal, T. Trzci{\'n}ski}

\author{Witold Oleszkiewicz$^1$, Peter Kairouz$^2$, Karol Piczak$^1$, Ram Rajagopal$^2$ and Tomasz Trzci{\'n}ski$^{1,3}$ }

%Please write out author names in full in the paper, i.e. full given and family names. 
%If any authors have names that can be parsed into FirstName LastName in multiple ways, please include the correct parsing, in a comment to the volume editors:
%\index{Lastnames, Firstnames}
%(Do not uncomment it, because you may introduce extra index items if you do that...)

\institute{$^1$Warsaw University of Technology, $^2$Stanford University, $^3$Tooploox
\email{witold.oleszkiewicz@pw.edu.pl}}

\maketitle

%===========================================================
\begin{abstract}
State-of-the-art machine learning algorithms can be fooled by carefully crafted adversarial examples. As such, adversarial examples present a concrete problem in AI safety. In this work we turn the tables and ask the following question: can we harness the power of adversarial examples to {\it prevent malicious adversaries from learning identifying information} from data while allowing non-malicious entities to {\it benefit from the utility} of the same data? For instance, can we use adversarial examples to anonymize biometric dataset of faces while retaining usefulness of this data for other purposes, such as emotion recognition? To address this question, we propose a simple yet effective method, called \textit{Siamese Generative Adversarial Privatizer} (SGAP), that exploits the properties of a Siamese neural network to find discriminative features that convey identifying information. When coupled with a generative model, our approach is able to correctly locate and disguise identifying information, while minimally reducing the utility of the privatized dataset. Extensive evaluation on a biometric dataset of fingerprints and cartoon faces confirms usefulness of our simple yet effective method.
\end{abstract}

%===========================================================
\section{Introduction}

Large-scale datasets enable researchers to design and apply state-of-the-art machine learning algorithms that can solve progressively challenging problems. Unfortunately, most organizations release datasets rather reluctantly due to the excessive amounts of sensitive information about participating individuals.

Ensuring the privacy of subjects is done by removing all personally identifiable information (e.g. names or birthdates) -- this process, however, is not foolproof. Correlation and linkage attacks~\cite{Narayanan08,Harmanci2016} often identify an individual by combining anonymized data with personal information obtained from other sources. Several such cases have been presented in the past, e.g. deanonymization of users' viewing history that was published in the Netflix Prize competition~\cite{Narayanan08}, identifying subjects in medical studies based on fMRI imaging data~\cite{Finn2015}, and linking DNA profiles of anonymized participants with data from the Personal Genome Project~\cite{Sweeney2013}.

Typical approaches to countering the shortcomings of anonymization techniques leverage data randomization. While randomizing datasets with differential privacy~\cite{Dwork08} provides much stronger privacy guarantees, the utility of machine learning models trained on such randomized data is often significantly impaired~\cite{Raval2017,Hayes2017,Kairouz16}. We therefore believe that there is an ever increasing need for new privatization methods that preserve the value of the data while protecting the privacy of individuals.

The above privacy problem becomes critical when dealing with sensitive biometric and medical images. Several breakthrough applications of computer vision have been proposed in this domain: \cite{Glasser2016} used machine learning algorithms to parcellate human cerebral cortex, \cite{Rajpurkar2017} utilized convolutional networks to detect arrhythmia, and  \cite{Famm2013} used machine learning to realize a precision medicine system. These applications, though critical for the advancement of the domain, rely on the access to highly sensitive data. This calls for novel privatization schemes that allow for the publication of images containing medical and biometric information without sacrificing the utility of the applications discussed above. 

\begin{figure}[bt!]
  \centering
  \includegraphics[width=0.55\textwidth]{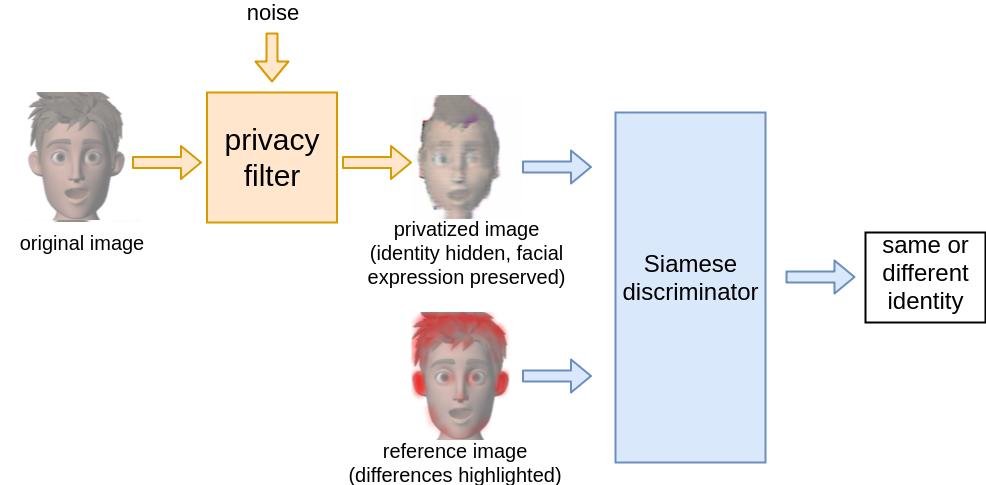}
    \caption{Basic functionality of the proposed Siamese Generative Adversarial Privatizer: given an original face image, the privacy filter generates a privatized image. The original identity is hidden, at the same time other useful features, e.g. facial expression, are preserved. Siamese discriminator identifies the discriminative features of the images.}
  \label{fig:teaser}
\end{figure}

\subsection{Our contributions}

In this work, we take a new approach towards enabling private data publishing. Instead of adopting worst case, context-free notions of statistical data privacy (such as differential privacy), we present a novel framework that allows the publisher to privatize images in a context-aware manner (Fig. \ref{fig:teaser}). Our framework builds up on the recent work~\cite{Huang17} where they propose a Generative Adversarial Privacy (GAP) method that casts the privatization as a constrained minimax game between a privatizer and an adversary that tries to infer private data. The approach we propose here is focused on biometric images and exploits a Siamese neural network architecture to identify image parts  that bear the highest discriminative power and perturb them to enforce privatization. Contrary to other works that quantify privacy in a subjective manner using user surveys~\cite{Orekondy_2018_CVPR}, we define here empirical conditions our privatizer needs to fulfill and propose metrics that allow to evaluate the privacy-utility trade-off we aim to explore. Finally, we present the results of our experiments on datasets of fingerprints and cartoon faces. Our results show that the proposed framework prevents an attacker from re-identifying privatized data while leaving other important image features intact. We call this approach \textit{Siamese Generative Adversarial Privatizer} (SGAP).

To summarize our contributions are twofold:
\begin{itemize}
\item a novel {\it privatization method} that uses a Siamese architecture to identify identity-discriminative image parts and perturbates them to protect privacy, while preserving the utility of the resulting data for other machine learning tasks, and
\item an empirical data-driven {\it privacy metric} (c.f. Section \ref{subsec:evaluation_metrics}) based on mutual information that allows to quantify the privatization effects on biometric images.
\end{itemize}

\subsection{Paper outline}
The remainder of this paper is organized as follows. In Sec.~\ref{sec:related}, we provide a brief survey of recent relevant works. In Sec.~\ref{sec:method}, we present the architectural details of our SGAP model. The main results of our paper are presented in  Sec.~\ref{sec:results}.  We~conclude our paper in Sec.~\ref{sec:conclusion}.

\section{Related Work}
\label{sec:related}

Privatization of data has been an active area of research with multiple works touching on this subject~\cite{Kairouz16,Raval2017,Hayes2017,Abadi17}. Our approach extends the concept of context-independent data privatization by incorporating context-dependent information as an input to the privatization algorithm. More precisely, it identifies the discriminative characteristics of the data and distorts them to enforce privacy. Although standard methods of protecting privacy based on erasing personal information have been widely used, correlation and linkage attacks allow to re-identify the users, even when explicitly identifying information is not present in the released datasets~\cite{Narayanan08}.

Those kinds of attacks pose an even greater threat to individual privacy when used against publicly available medical databases~\cite{Gymrek2013}. \cite{Harmanci2016} show that using publicly available genotype-phenotype correlations, an attacker can statistically relate genotype to phenotype and therefore re-identify individuals. Publicly available profiles in the Personal Genome Project can be linked with names by using demographic data \cite{Sweeney2013}. Also, when considering fRMI imaging data, individual variability across individuals is both robust and reliable, thus can be used to identify single subjects \cite{Finn2015}.

Although numerous works are focused on finding discriminative patterns within the data~\cite{Fisher36,Trzcinski2012}, we use a Siamese neural network architecture~\cite{Bromley94} since it allows us to learn a discriminant data embedding in an end-to-end fashion. Contrary to the typical goal of a Siamese architecture, i.e. learning similarity, we use it to identify discriminant parts of a pair of images and alter those parts with minimal impact on other useful features. When both examples come from the same individual, this setup allows us to learn a perturbation that carefully disguises the individual's identity, hence protecting their privacy.

One can consider the problem of data anonymization to be conceptually similar to the idea of adversarial examples in neural network architectures~\cite{Liang17,Kos17,Lee17,Baluja17,Tripathy17}. In the case of adversarial examples, the adversary wants to trick the neural network into misclassifying a slightly perturbed input of a given class. Similarly, our goal is to modify the data points in such a way that the identity of the individual corresponding to the data cannot be correctly classified. The most relevant work is~\cite{Lee17}, where they use a~Generative Adversarial Network (GAN)~\cite{Goodfellow14} framework to create adversarial examples and use them in training to increase the robustness of the classifier.

Similar to us, \cite{Orekondy_2018_CVPR} analyses the trade-off between data privacy and utility. In their work, however, privacy and utility metrics are defined based on a user-study, where the users were asked to assess the usefulness of the anonymized images in the context of social media distribution. The privacy, on the other hand, was measured by first enlisting a number of attributes linked to privacy ({\it e.g.} passport number or registration plates) and then asking the users to validate if a given privacy attribute is visible in the photo or not. We argue that this way of measuring both privacy and utility is limited to a very specific subset of applications. In our work we propose fundamentally different metrics for both privacy and utility that have backing in information theory and machine learning.

Another relevant and recent works~\cite{Tripathy17},~\cite{Chen2018} address the privatization problem using a generative adversarial approach while providing theoretical privacy-utility trade-offs. The work of \cite{Chen2018}, which is the most similar to our work, proposes an architecture combining Variational Autoencoder (VAE) and GAN to create an identity-invariant representation of a face image. Their approach differs from ours as they use an additional discriminator, which explicitly controls which useful features of the images are to be preserved, whereas in our approach the model has no information about other features of the images, except that it knows whether a pair of images belongs to the same person or different people. This is a significant contribution because in practice, one cannot expect to know all potential applications of the privatized images. Therefore our approach proves to be more robust towards real-life applications.

\cite{Oh2017} presents a similar game-theoretic perspective on image anonymization. However, the difference is that it focuses on adversarial image perturbations (carefully crafted perturbations invisible to human), while our privatizer introduces structural changes to the image. In \cite{Sun2017}, a head inpainting obfuscation technique is proposed by generating a realistic head inpainting using facial landmarks. On contrary, our goal is to hide the identity of a person without knowing which part of the image is responsible for identity. Thanks to this, our framework is more universal and has a much wider field of application, not only to hide face identity, but also hide identity in cases where there is no prior knowledge of which part of the image should be obfuscated. \cite{Mirjalili2018} and \cite{Mirjalili2017} are relevant to our work and deal with a problem similar to ours. However, the formulation of the problem is different from ours. \cite{Mirjalili2018} and \cite{Mirjalili2017} transform an input face image in a way such that the transformed image can be successfully used for face recognition (so the identity is preserved) but not for gender classification. Our goal is the opposite, we want to hide identity while maintaining as much other features as possible, without explicitly modeling the non-malicious classification tasks. Another difference is that our model requires only identity labels. The architecture of the models presented in \cite{Mirjalili2018} and our work are similar, however we use Siamese discriminator what makes our approach advantageous when applied to large datasets with thousands or even millions of people, since this architecture reduces the output of the discriminator to a binary output rather than create a~long list of individual class predictions.

\section{Method}
\label{sec:method}

The goal of our approach is to develop a privatizer that converts an input image into its privatized version in such a way that: (1) the privacy of the subject is preserved by making sure that the identifying features are hidden, (2) the utility of the original image is maintained by preserving the non-identifying features that are vital for other machine learning tasks, and (3) the privacy-utility trade-off can be adjusted.

\subsection{Proposed approach}
\noindent To enforce the above conditions, we will use a custom neural network architecture, dubbed \textit{Siamese Generative Adversarial Privatizer}, that consists of two tightly coupled models: a generator $G(\theta_g)$ and a discriminator $D(\theta_d)$. This coupling is inspired by Generative Adversarial Networks (GANs)~\cite{Goodfellow14}. Two neural networks compete with each other: the discriminator tries to predict the identity of the person in the image, while the generator tries to generate such an image which fools the discriminator and thus hides the identity of the person.

We use a Siamese architecture~\cite{Bromley94} for the discriminator. This allows us to extract discriminative and identifying features from images. More importantly, this architecture reduces the output of the discriminator to a single value (from 0 to 1) rather than create a~long list of individual class predictions, an approach which would be prohibitive when applied to large datasets with thousands or even millions of people. In this case, we use pairs of images (instead of single images) to train the neural network, and the goal of the Siamese discriminator is to classify whether the two images belong to the same person or to different people.

Furthermore, the above problem is subjected to a distortion constraint, which ensures that the privatized images are not too different from the original images. 

\begin{figure}[bt!]
  \centering
  \includegraphics[width=0.6\textwidth]{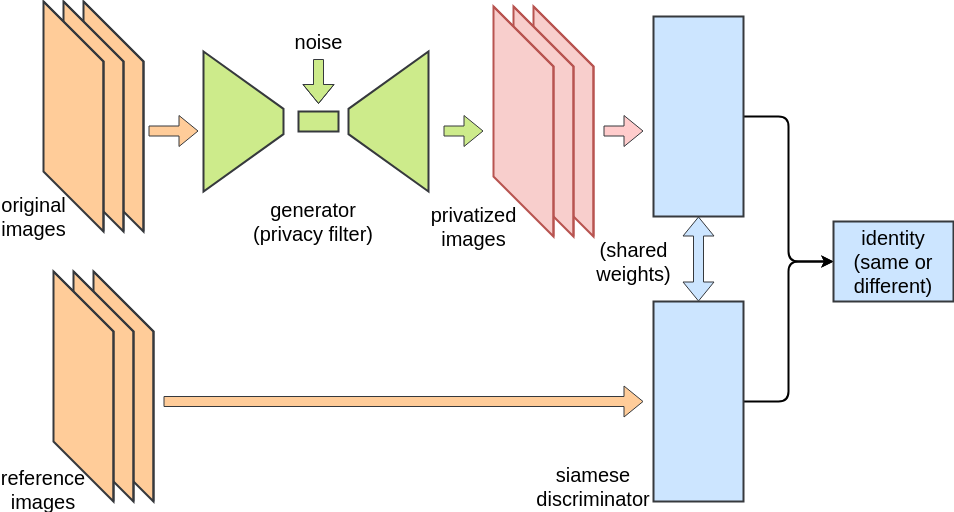}
  \caption{Overview of our Siamese Generative Adversarial Privatizer model. The generator acts as a privacy filter, which hides the identity of the person in the original images. The Siamese discriminator recognizes whether the person in the privatized image is the same person as in the reference image.}
  \label{fig:model}
\end{figure}

We did not use $L_2$ since it is sensitive to small changes (e.g. shift, rotation, etc.) which do not significantly affect the content of the image. Instead we chose SSIM (structural similarity index) \cite{Wang2004} which is sensitive to the structural changes of images, not pixel-by-pixel differences like $L_2$ ~\cite{DBLP:journals/corr/ZhaoGFK15}. We enforce a constraint on SSIM which allows us to control the level of distortion added to protect identity, and thus ensure that the quality of privatized images is not substantially degraded. The architecture overview can be seen in Fig.~\ref{fig:model}.

\subsection{Architecture}

\begin{figure}[bt!]
  \centering
  \includegraphics[width=0.6\textwidth]{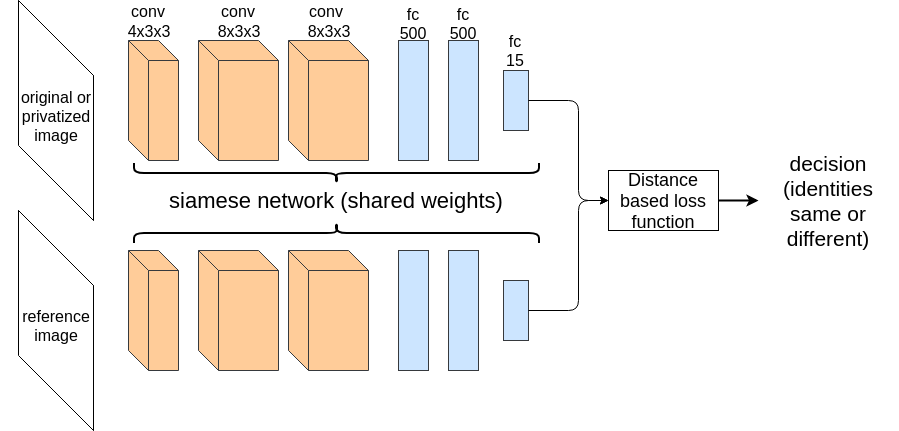}
  \caption{Discriminator's architecture. We use a Siamese neural network to verify the identities of people in the images. The discriminator classifies whether a pair of images belongs to the same person or to different people. We get the output from the range between 0~and~1 applying distance-based loss function to the output of the last fully connected layer of the Siamese discriminator.}
  \label{fig:discriminator}
\end{figure}

Our discriminator is a Siamese convolutional neural network, which consists of two identical branches with shared weights, as shown in Fig.~\ref{fig:discriminator}. Each branch consists of 3 blocks of the following form: (1) Convolutional layer (mask $3\times3$, stride=1, padding=0), (2) Leaky rectified linear unit ($\alpha=0.1$), (3) Batch normalization, (4) Dropout ($p=0.2$). The blocks are followed by 2 dense layers (500 neurons, leaky rectified linear unit, $\alpha=0.1$) and an output layer (15 neurons). A discriminator network converts two input images to two output representations (embeddings) $D(\mat{X}_1, \mat{X}_2)  \rightarrow (\vec{o}_1, \vec{o}_2)$.

\begin{figure}[bt!]
  \centering
  \includegraphics[width=0.6\textwidth]{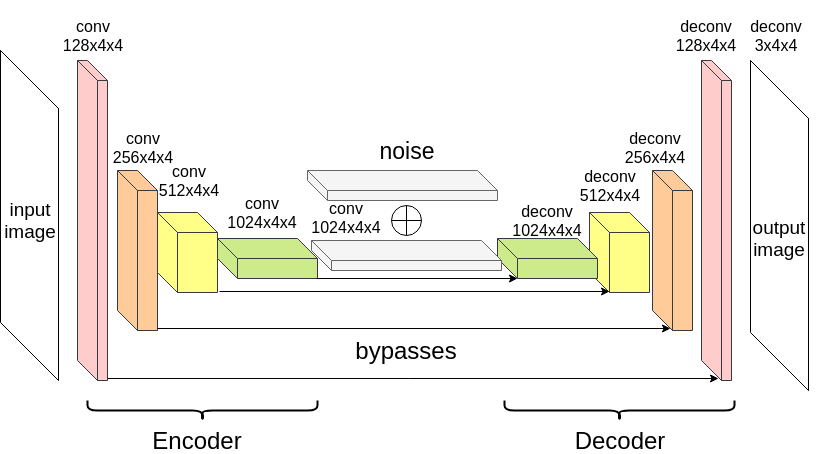}
    \caption{Generator's architecture. We use Variational Autoencoder-like architecture to generate a privatized image in a context-aware manner based on the original image. At the bottleneck of the generator we get a compressed representation of the image without identity features, and thanks to the bypasses between the layers we preserve other useful features of the original image.}
  \label{fig:generator}
\end{figure}

The generator network, as presented in Fig.~\ref{fig:generator}, consists of two parts: the encoding part and the decoding part. The encoder follows the typical architecture of a convolutional neural network. It consists of 5 blocks of the following form: (1) Convolutional layer (mask $4\times4$, stride=2, padding=1), (2) Leaky rectified linear unit ($\alpha=0.1$), (3) Batch normalization. At each downsampling step we double the number of feature
channels.

The decoder consists of 5 blocks of the following form: (1) Transpose convolutional layer (mask $4\times4$, stride=2, padding=1), (2) Leaky rectified linear unit ($\alpha=0.1$),  (3) Batch normalization, (4) Dropout ($p=0.5$). At each upsampling step we halve the number of feature channels. Also we concatenate the feature maps of the decoder part with the corresponding feature map from the encoder part (these are bypasses). Last deconvolutional layer is followed by a hyperbolic tangent activation function.

A noise matrix $\randmat{Z}$ is added to the bottleneck part of the generator, {\it i.e.} to the latent space variable representing input image in a low-dimensional space. We use a noise matrix instead of a vector, as we do not use a standard fully-connected layers in our generator and retain convolutional layers instead. The output of generator network is a privatized version of original image: $G(\randmat{Z}, \mat{I}) \rightarrow \tilde{\mat{I}}$.

\subsection{Training}

When iterating over training dataset we get tuples: $(\mat{I}_i, \mat{I}^{'}_i, l_i)$, where $\mat{I}_i$ and $\mat{I}^{'}_i$ is a pair of images and $l_i$ is a binary label where $l_i = 0$ if the images have the same identity and $l_i = 1$ for different identities. There are two types of pairs in the training set. Firstly, when the generator is turned off, $\mat{I}_i, \mat{I}^{'}_i$ are images from the original training set. Secondly, when the generator is turned on, $\tilde{\mat{I}_i}=G(\randmat{Z}_i, \mat{I}_i)$ is the privatized version of the image $\mat{I}_i$ from the original training set. $\mat{I}^{'}_i$ is the reference image, also from the original training set. In both cases mentioned above we use stratified random sampling in order to balance two classes: $l = 0$ and $l = 1$.

The discriminator $D$ takes a pair of images $\mat{I}, \mat{I}^{'}$ and outputs a probability that both images come from the same person, i.e. $l = 0$, based on a distance-based metric:

\begin{equation*}
D(\mat{I}, \mat{I}^{'}) \rightarrow \frac{1 + e^{-m}}{1 + e^{d(\vec{o}, \vec{o}^{'})^2-m}} = P(\mat{I} \stackrel{\mathclap{\scriptsize\mbox{sim.}}}{\sim} \,\mat{I}^{'})
\end{equation*}

\noindent where $m$ is a predefined margin and $d(\vec{o}, \vec{o}^{'})$ is an Euclidean distance between embeddings $\vec{o}$ and $\vec{o}^{'}$ in the last fully connected layer of the discriminator. Given this formulation of the discriminator we use a cross entropy loss for training: 
\begin{equation*}\mathcal{L}(l, D(\mat{I}, \mat{I}^{'})) = -(1 - l) \log D(\mat{I}, \mat{I}^{'}) - l \log\left(1 - D(\mat{I}, \mat{I}^{'})\right)
\end{equation*}

We train our model similarly to GAN. When the generator training is frozen, our goal is to train the discriminator to recognize whether a pair of images belongs to the same person or to different people. When the generator is trained, there is a minmax game between the generator and the discriminator in which the generator is trying to fool the discriminator and generate an image that hides the identity of the subject. The training equation for our privatization task is:

\begin{align*}
\min_D \max_G \frac{1}{N} \sum_{i=0}^{N-1} \mathcal{L}(l_{i}, D(\mat{I}_i, \mat{I}^{'}_i)) + \frac{1}{N} \sum_{i=0}^{N-1} \mathcal{L}(0, D(\mat{I}^{'}_i, G(\randmat{Z}_i, \mat{I}_i)))
\end{align*}

\noindent Furthermore, the above minimax optimization problem is subject to the following critical constraint: $\frac{1}{N} \sum_{i=0}^{N-1} d(\mat{I}_i, G(\randmat{Z}_i, \mat{I}_{i})) < \delta,$
where $d(x,y)$ is a distortion metric and $\delta$ is a distortion threshold. The distortion constraint is used to limit all the other image changes except for hiding identity and therefore the utility of the images is preserved. We use Structural Similarity Index as the distortion metric. The above constraint can be incorporated into the main minimax objective function as follows: 
\begin{equation}
\min_D \max_G \sum_{i=0}^{N-1} \mathcal{L}(l_{i}, D(\mat{I}_i, \mat{I}^{'}_i)) +  \sum_{i=0}^{N-1} \mathcal{L}(0, D(\mat{I}^{'}_i, G(\randmat{Z}_i, \mat{I}_i)))  +  \lambda  \sum_{i=0}^{N-1} d(\mat{I}_i, G(\randmat{Z}_i, \mat{I}_{i}))
\label{eq:loss}
\end{equation}

Our Siamese Generative Adversarial Privatizer network is trained for 100 epochs using ADAM optimizer with $\beta_1=0.9$ and $\beta_2=0.999$.

\section{Results}
\label{sec:results}
In this section, we present the results of evaluation of our method. We first present the datasets and evaluation metrics. Then we show qualitative and quantitative results of our evaluation that confirm usefulness of our approach in the context of data privatization.

\subsection{Datasets}
\subsubsection{Fingerprints}
To validate how well our method performs in terms of identity privatization, we evaluate it on a dataset of fingerprints. Although the main purpose of fingerprint datasets is to identify people and therefore their privatization may not be needed in their real-life use cases, we treat this dataset as our toy example and evaluate how well we can hide the privacy of the fingerprint owner. Since there exists a trade-off between the privatization and the utility of the resulting data, we refer to a proxy task of finger type classification to validate how useful our privatization method is. In other words, we try to classify the type of the finger ({\it e.g.} middle finger, index finger, ring finger) while gradually increasing the privacy of the dataset. Sec.~\ref{subsec:quantitative} presents the results of this experiment.

We use NIST 8-Bit Gray Scale Images of Fingerprint Image Groups \cite{NIST}. This database contains 4000 8-bit grayscale fingerprint images paired in couples. Each image is 512-by-512 pixels with 32 rows of white space at the bottom. We use only one image of each pair in our experiments. The dataset contains images for 2000 individuals. For each person there are two different fingerprint shots of the same finger (denoted as: $f$, $s$). Our method requires pairs of images as input. In each epoch the dataset is iterated over 4000 pairs of images. 

For the first half of the pairs when index of a pair is $\mathit{i} < 2000$ we return a label $\mathit{l} = 0$ and a pair of images ($f$, $s$) belonging to the person with $\mathit{ID} = i$.

For the second half of the pairs when index $\mathit{i} >= 2000$ we return a label $\mathit{l} = 1$ and two images. First image is image $f$ of person with $\mathit{ID} = i - 2000$. Second image is an image ($f$ or $s$) of a different person (selected at random).

This way we have a 50\%/50\% split over similar/dissimilar pairs and the dataset loader is quasi-deterministic (for a given index $\mathit{i}$ the first image is guaranteed to be constant).

\subsubsection{Animated faces}

The second dataset that we use is FERG dataset \cite{Deepali2017}. FERG is a dataset of cartoon characters with annotated facial expressions. It contains 55769 annotated face images of six characters. The images for each identity are grouped into 7 types of facial expressions, such as: anger, disgust, fear, joy, neutral, sadness and surprise.

In each epoch the dataset is iterated over 10000 pairs of images. For the first half of the pairs we use different randomly selected images of the same person. In this case $\mathit{l} = 0$. For the second half of the pairs we use randomly selected images of different people. In this case $\mathit{l} = 1$.
This way we have a 50\%/50\% split over similar/dissimilar pairs and the dataset loader is quasi-deterministic.

\subsection{Evaluation metrics}
\label{subsec:evaluation_metrics}
To evaluate the performance of our SGAP model and show that it learns privacy schemes that are capable of hiding biometric information even from computationally unbounded adversaries, we propose computing the mutual information between: (a) $X = (X_1, X_2)$ where $X_1$ is a privatized image and $X_2$ is an original image, and (b) $Y$ where $Y=0$ when $X_1$ and $X_2$ belong to the same person and $Y=1$ when they belong to different people. $X_1$ is privatized using the scheme that is learned in a data-driven fashion using SGAP. By Fano's inequality, if $I(X;Y)$ is low then $Y$ cannot be learned from $X$ reliably (even under computationally infinite adversaries) \cite{Cover2006}. In other words, if $I(X;Y)$ is sufficiently small, there's no way we can reliably learn whether or not a privatized image belongs to the same person in another non-privatized image. This ensures that privacy is guaranteed in a strong sense.

In practice, we do not have access to the joint distribution $P(X,Y)$. We instead have access to a dataset of i.i.d observations $\mathcal{D} = \{(X_i, Y_i\}_{i=1}^{n} \}$. Here, the $X_i$'s are computed after the SGAP training phase is over by applying the learned privacy scheme on a separate test set. We are thus interested in empirically estimating $I(X;Y)$ from $\mathcal{D}$. We will call this estimate ``empirical mutual information'' and denote it by $\hat{I}(X; Y)$. To compute $\hat{I}(X; Y)$, we can use the following formula: 
\begin{align*}
\hat{I}(X; Y) = \hat{H}(X) - \hat{H}(X|Y) 
\end{align*}
where $\hat{H}(X)$ and $\hat{H}(X|Y)$ are the empirical entropies of $X$ and $X$ given $Y$. To compute these empirical entropies, we use the Kozachenko-Leonenko entropy estimator \cite{Fournier2016} which we briefly explain next. Letting $R_i = \min_{j, j \neq i }\|X_i - X_j\|$, for $j = 1, \dots, n$, we get
\begin{align*}
	\hat{H}(X) & = \frac{1}{n} \sum_{i=1}^n \log \big( (n-1)R_i^{d} \big) + constant \\
    & = \frac{d}{n}\sum_{i=1}^n \log R_i + \frac{1}{n} \sum_{i=1}^n \log(n-1) + constant
\end{align*}
where $d$ is the dimension of $X$, i.e. $X_i \in \mathbb{R}^{d}$. Assuming we have a two-class problem ($Y=0$ for same identities, $Y=1$ for different identities), the conditional entropy is given by
\begin{align*}
	\hat{H}(X|Y) = \hat{H}(X|Y=0)\hat{P}(Y=0) + \hat{H}(X|Y=1)\hat{P}(Y=1)    
\end{align*}
Notice that $\hat{P}(Y=0)= \frac{n_0}{n}$, $\hat{P}(Y=1)= \frac{n_1}{n}$, where $n_0$ and $n_1$ are the counts of samples with label $Y$ equals 0 and 1 respectively. We divide sample $X$ into two partitions. Letting $i_1, i_2, \dots, i_{n_0} $ be the indices corresponding to $Y_i = 0$, we have a set $\mathcal{X}_0 = \{X_{i_1}, X_{i_2}, \dots, X_{i_{n_0}}\}$. Automatically we have ${i^\prime}_1, i^{\prime}_2, \dots, i^{\prime}_{n_0} $, the indices of samples associated with $Y_i=1$. Thus, we get  $\mathcal{X}_1 = \{X_{i^{\prime}_1}, X_{i^{\prime}_2}, \dots, X_{i^{\prime}_{n_1} } \}$. Therefore we calculate the nearest neighbor distance for each sample within the particular set as follows:

\begin{align*}
	R_{i_k} = \min_{l\neq k, l=1,\dots, n_0 } \|X_{i_k} - X_{i_l}\| \quad \quad
    R_{i^{\prime}_k} = \min_{l\neq k, l=1,\dots, n_1} \|X_{i^{\prime}_k} - X_{i^{\prime}_l}\|
\end{align*}

\begin{align*}
    \hat{H}(X|Y=0) = \frac{1}{n_0} \sum_{k=1}^{n_0} \log\big( (n_0-1) R_{i_k}^d \big) + constant \\
    \hat{H}(X|Y=1) = \frac{1}{n_1} \sum_{k=1}^{n_1} \log\big( (n_1-1) R_{i^{\prime}_k}^d \big) + constant 
\end{align*}

Then the empirical mutual information can be expressed as 
\begin{align*}
	& \hat{I}(X,Y)  = \hat{H}(X) - \Big( \hat{H}(X|Y=0)\hat{P}(Y=0) + \hat{H}(X|Y=1)\hat{P}(Y=1) \Big) \label{Privacy:MI:eq:emi_1} \\
    & = \frac{1}{n} \sum_{i=1}^n \log\big((n-1) R_{i}^d \big) + \nonumber \\ & - \Bigg(\Big( \frac{1}{n_0} \sum_{k=1}^{n_0} \log\big( (n_0-1) R_{i_k}^d \big)\Big) \frac{n_0}{n} + \Big( \frac{1}{n_1} \sum_{k=1}^{n_1} \log\big( (n_1-1) R_{i^{\prime}_k}^d \big) \Big) \frac{n_1}{n}   \Bigg) %\label{Privacy:MI:eq:emi_2}
\end{align*}

To estimate values of $R_{i_k}$ and $R_{i^{\prime}_k}$ we use $L_2$ norm between image pixels projected to a 3-dimensional space via t-SNE~\cite{vanDerMaaten2008}. We reduce the dimensionality to increase the efficiency of computation, but our metric remains agnostic to image distance calculation and other methods can also be used here.% can also be calculate%projected  {ie need to calculate distance between the images to estimate entropy. Although several metrics can be used here, {\it e.g.} $L_2$ norm, we refer to a more data-dependent method and calc As a distance between pairs of images we take a t-SNE-projected input dimensions. High dimensional pair of images is reduced to a point in 3-dimensional space using initially PCA (reduction to 50 dimensions), then t-SNE algorithm (reduction to 3 dimensions). 

The second approach to quantify privacy is by measuring an identity misclassification rate. We measure what percentage of privatized images effectively fool our Siamese discriminator.

To quantify utility of privatized dataset we measure accuracy of the proxy classification task (finger type classification for fingerprint dataset and facial expression classification for faces dataset). More precisely, we evaluate how good in terms of accuracy a separate independent method can be trained for using a privatized dataset. We use fine-tuned ResNet architecture, pre-trained on ImageNet without freezing. In addition we split the dataset into training and validation. The accuracy is measured using k-fold validation ($k=4$).

\subsection{Qualitative results}

In this section, we present the qualitative results of our evaluation, demonstrating the ability of our network to increase the privacy of input data.

\begin{figure}[t!]
\centering
\includegraphics[width=0.6\textwidth]{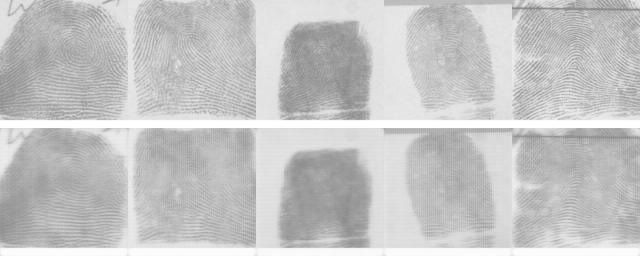}
\hbox{\hspace{0.1em} \includegraphics[width=0.6\textwidth]{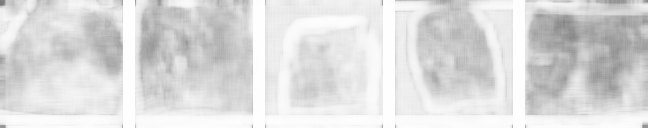}}

\caption{A toy example of how our privatization method can hide identities of the fingerprint owners. Original fingerprints in the upper row. Fingerprints with added artifacts that fool identity discriminator in the middle row. Structural Similarity difference~\cite{Wang2004} of the original and privatized images is presented in the bottom row. Our Siamese Generative Adversarial Privatizer learns to locate discriminant image features, such as fingerprint minutiae, and substitutes them with anonymizing artifacts. Although in practice fingerprints are used for person identification, we validate if privatized images can be useful ({\it i.e.} if they can retain utility) for a proxy task of finger type classification. Since our method does not add noise arbitrarily across the image, but only focuses on hiding sensitive personal information, the resulting dataset can be published and used by machine learning for other tasks, e.g. finger type classification or skin disease detection.}
\label{fig:fingerprints}
\end{figure}

\begin{figure}[bt!]
\centering
\includegraphics[width=0.6\textwidth]{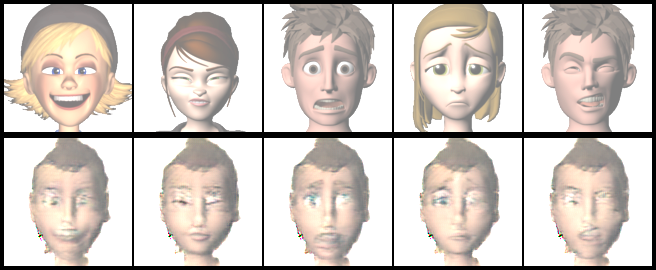}
\caption{Original cartoon faces in the upper row. Privatized versions of cartoon faces in the bottom row. Our Siamese Generative Adversarial Privatizer learns to hide the identity of the people, while other important image features, such as facial expression remain intact.}
\label{fig:cartoons}
\end{figure}

\begin{figure}[bt!]
\centering
\includegraphics[width=0.6\textwidth]{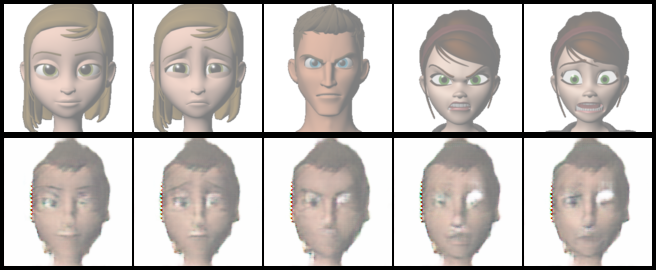}
\caption{Too much privacy, utility is not preserved. Original cartoon faces in the upper row. Privatized versions of cartoon faces in the bottom row. Our model has been tuned too much towards ensuring privacy, so that the utility of the images has not been preserved, facial expressions are hard to recognize.}
\label{fig:cartoons_too_much_privacy}
\end{figure}

\begin{figure}[bt!]
\centering
\includegraphics[width=0.6\textwidth]{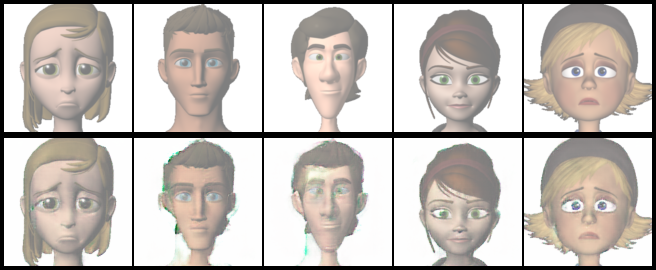}
\caption{Not enough privacy, utility is preserved. Original cartoon faces in the upper row. Privatized versions of cartoon faces in the bottom row. Our model has been tuned too much towards preserving utility, so that the identities of the people in the images are not hidden, only minor changes have been added to the images.}
\label{fig:cartoons_too_much_utility}
\end{figure}

\begin{figure}[bt!]
\centering
\includegraphics[width=0.75\textwidth]{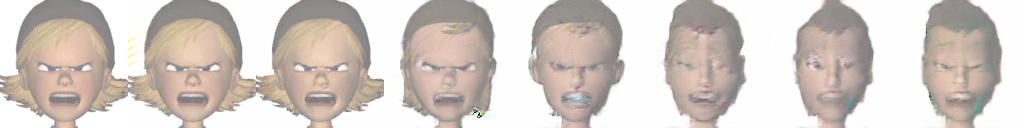}
\includegraphics[width=0.75\textwidth]{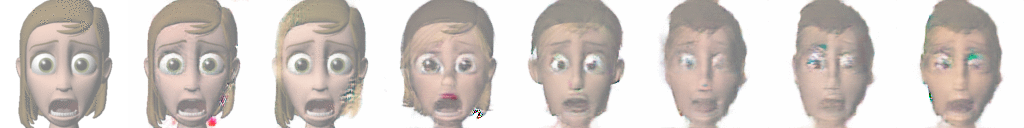}
\caption{Images in the first column are the original ones, next there are privatized images generated for different values of distortion constraint $\lambda \in \{10, 8, 6, 4, 2, 1, 0.7\}$. Original images of different identities collapse into an anonymous identity with the expression preserved from the original image.}
\label{fig:lambda}
\end{figure}

Fig.~\ref{fig:fingerprints} and~\ref{fig:cartoons} show sample results obtained as an output of our privatization. In Fig.~\ref{fig:cartoons} we see that the identities of people have been hidden, while other useful features, in this case facial expressions, have been preserved.  Fig.~\ref{fig:cartoons_too_much_privacy},~\ref{fig:cartoons_too_much_utility} and~\ref{fig:lambda} illustrate the trade-off between utility and privacy while tuning $\lambda$ distortion metric constraint. We see that by tuning the $\lambda$ parameter we can adjust the level of privacy and utility, finally finding the optimal value for both privacy and utility.

\subsection{Quantitative results}
\label{subsec:quantitative}
To obtain quantitative results we train our SGAP model with different values of maximal distortion constraint $\lambda$ (see Eq.~\ref{eq:loss}) in order to adjust the privacy level of the dataset. The goal of our generator is to add such noise to the latent space that privatized image fools the discriminator, which the discriminator in turn has to verify if the pair of images comes from the same person. After SGAP is trained, the generator part can be used to privatize datasets.

To measure the utility of the privatized fingerprints dataset, we refer to a proxy task of finger type classification. Although in fingerprints are typically used to identify the identity of an individual, in our case we use the proposed privatization method to hide the identity and anonymize the dataset. The objective of this experiment is to evaluate how increasing data privacy effects the utility of the resulting dataset when used as training data for a machine learning algorithm. Hence, we use a proxy machine learning task, finger type classification. To measure the utility of the privatized cartoon faces dataset, we use facial expression classification as machine learning task. As a classifier, trained on privatized datasets, we use fine-tuned ResNet architecture, pre-trained on ImageNet without freezing. For each dataset generated using different maximal distortion constraint, we calculate classification accuracy and quantify the privacy by estimation of mutual information (fingerprint dataset) or using identity misclassification rate (faces dataset). 

\begin{figure}[b!]
\centering
\begin{minipage}{.48\textwidth}
  \centering
  \includegraphics[width=1.0\linewidth]{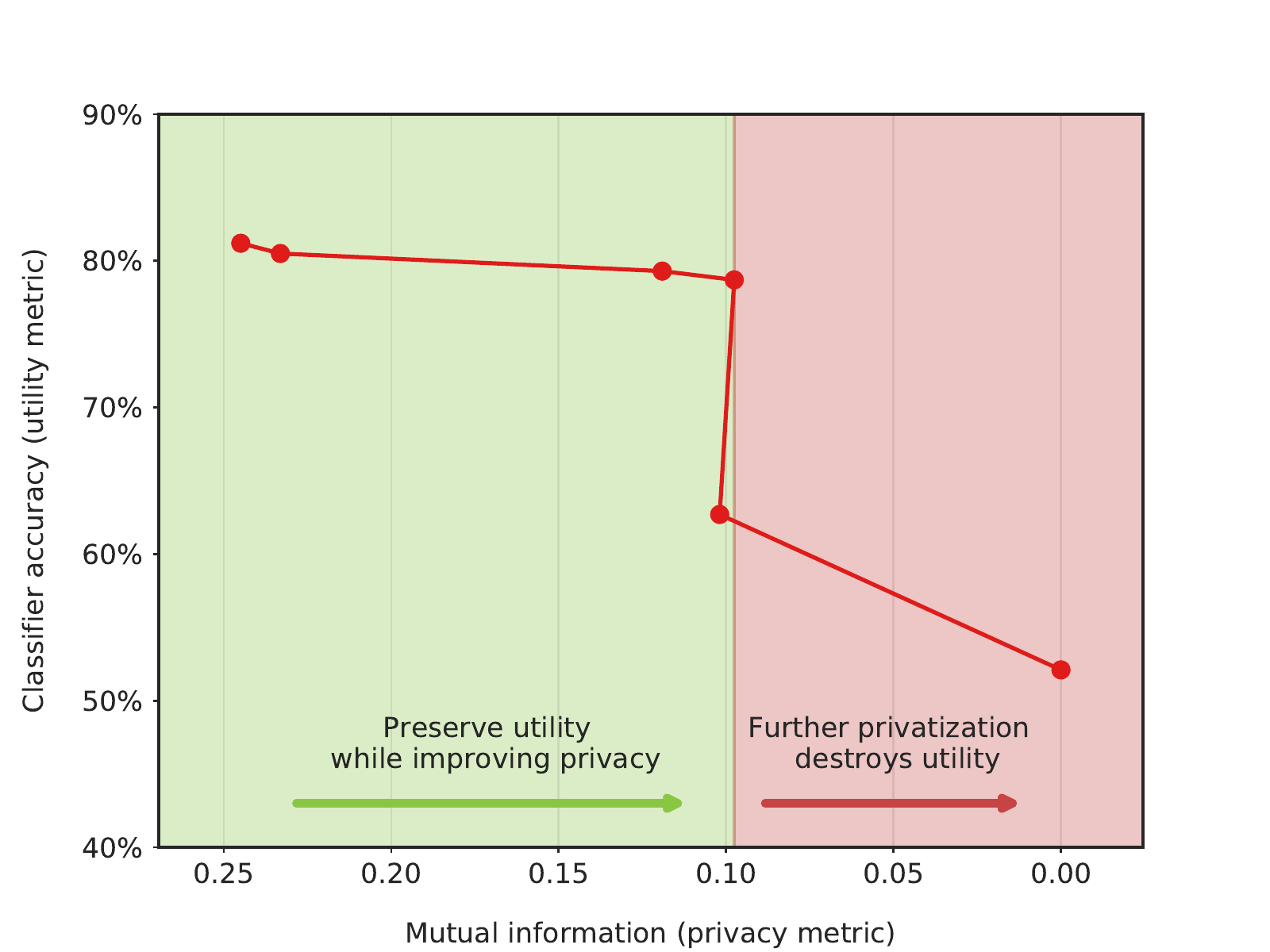}
  \caption{Graph of mutual information estimation and the accuracy of a classifier trained with fingerprint dataset privatized with different maximal constraint distortion thresholds. In green the region where the utility of dataset is preserved while the likelihood of classifying the privatized version of the image as belonging to a given person is reduced. This result proves that by using our privatization method we are able to significantly increase the privacy of the biometric dataset, while not reducing its utility for a proxy task of finger type classification.}
  \label{fig:mi_acc}
\end{minipage}%
\hfill
\begin{minipage}{.48\textwidth}
  \centering
\includegraphics[width=1.0\linewidth]{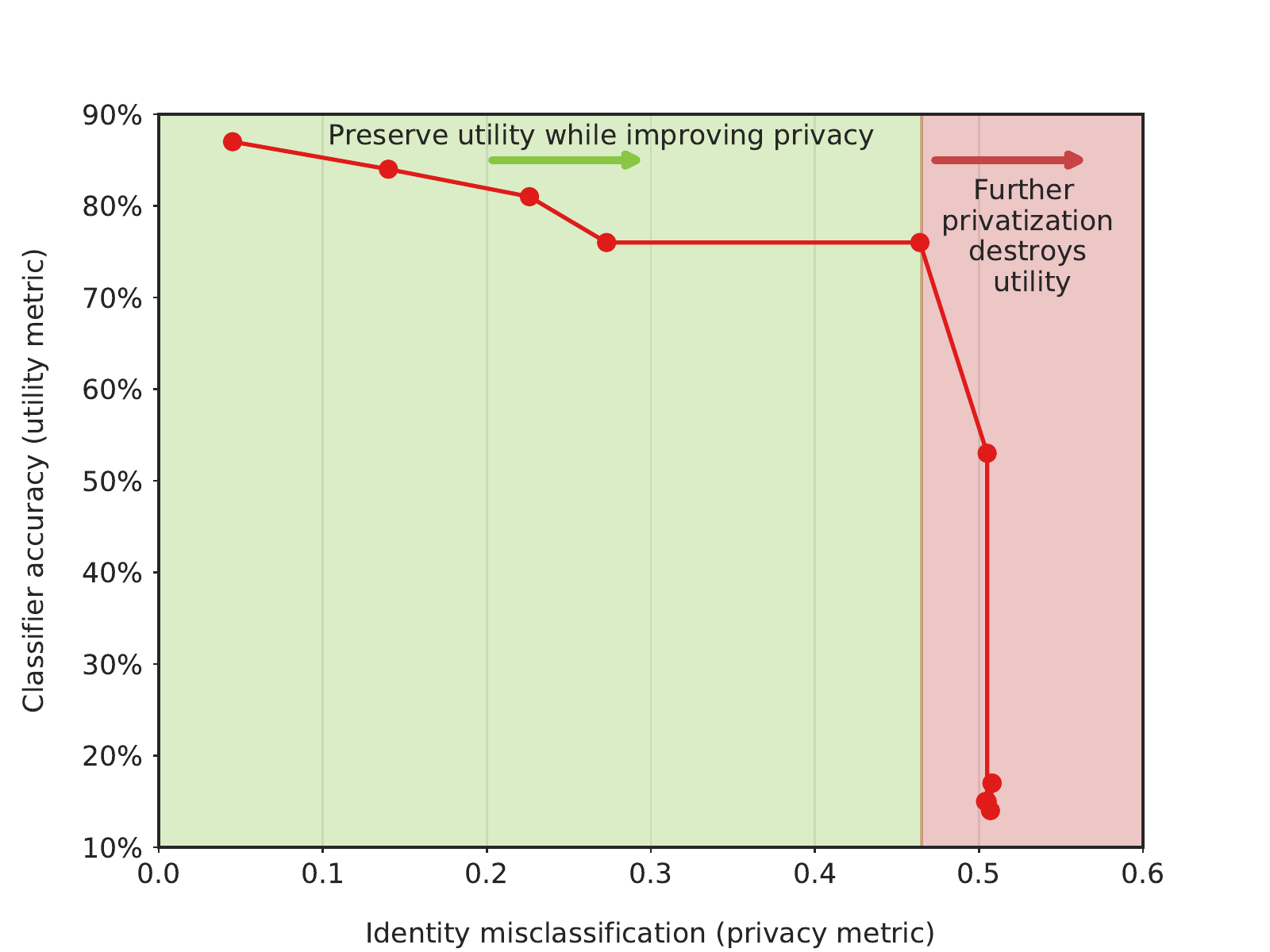}  \caption{Graph of identity misclassification rate and the accuracy of a classifier trained with cartoon faces dataset privatized with different maximal constraint distortion thresholds. In green the region where the utility of dataset is preserved while the likelihood of classifying the privatized version of the image as belonging to a given person is reduced. This result proves that by using our privatization method we are able to significantly increase the privacy of the biometric dataset, while not reducing its utility for a task of facial expression classification.}
  \label{fig:mi_acc_faces}
\end{minipage}
\end{figure}

Fig.~\ref{fig:mi_acc}~and~\ref{fig:mi_acc_faces} show the results. In both cases we see a significant drop in privacy metric, while for the same range of parameters, the accuracy of the classifier remains stable, indicating that the utility of the dataset is not decreased.

\section{Conclusions}
\label{sec:conclusion}

We presented the \textit{Siamese Generative Adversarial Privatizer} (SGAP) model for privacy-preserving of biometric data. We proposed a novel architecture combining Siamese neural network, autoencoder, and Generative Adversarial Network to create a context-aware privatizer. Experimental results on two public  datasets demonstrate that our approach strikes a balance between privacy preservation and dataset utility.

\section*{Acknowledgment}

The work was partially supported as RENOIR Project by the European Union Horizon 2020 research and innovation programme under the Marie Sk{\l}odowska-Curie grant agreement No 691152 (project RENOIR) and by Ministry of Science and Higher Education (Poland), grant No. W34/H2020/2016. We thank NVIDIA Corporation for donating Titan Xp GPU that was used for this research.

\bibliographystyle{splncs03}
\bibliography{egbib}

\begin{thebibliography}{10}
\providecommand{\url}[1]{\texttt{#1}}
\providecommand{\urlprefix}{URL }

\bibitem{Abadi17}
Abadi, M., Erlingsson, U., Goodfellow, I., McMahan, H.B., Mironov, I.,
  Papernot, N., Talwar, K., Zhang, L.: On the protection of private information
  in machine learning systems: Two recent approaches. CoRR  abs/1708.08022
  (2017)

\bibitem{Deepali2017}
Aneja, D., Colburn, A., Faigin, G., Shapiro, L., Mones, B.: Modeling stylized
  character expressions via deep learning. In: Lai, S.H., Lepetit, V., Nishino,
  K., Sato, Y. (eds.) Computer Vision -- ACCV 2016. pp. 136--153. Springer
  International Publishing, Cham (2017)

\bibitem{Baluja17}
Baluja, S., Fischer, I.: Adversarial transformation networks: Learning to
  generate adversarial examples. CoRR  abs/1703.09387 (2017)

\bibitem{Bromley94}
Bromley, J., Guyon, I., LeCun, Y., S\"{a}ckinger, E., Shah, R.: Signature
  verification using a "siamese" time delay neural network. In: Advances in
  Neural Information Processing Systems 6, pp. 737--744. Morgan-Kaufmann (1994)

\bibitem{Chen2018}
Chen, J., Konrad, J., Ishwar, P.: Vgan-based image representation learning for
  privacy-preserving facial expression recognition. CoRR  abs/1803.07100
  (2018), \url{http://arxiv.org/abs/1803.07100}

\bibitem{Cover2006}
Cover, T.M., Thomas, J.A.: Elements of Information Theory (Wiley Series in
  Telecommunications and Signal Processing). Wiley-Interscience, New York, NY,
  USA (2006)

\bibitem{Dwork08}
Dwork, C.: Differential privacy: A survey of results. In: International
  Conference on Theory and Applications of Models of Computation. pp. 1--19
  (2008)

\bibitem{Famm2013}
Famm, K., Litt, B., Tracey, K.J., Boyden, E.S., Slaoui, M.: Drug discovery: A
  jump-start for electroceuticals. Nature  496(7444),  159--161 (2013)

\bibitem{Finn2015}
Finn, E.S., Shen, X., Scheinost, D., Rosenberg, M.D., Huang, J., Chun, M.M.,
  Papademetris, X., Constable, R.T.: Functional connectome fingerprinting:
  identifying individuals using patterns of brain connectivity. Nat Neurosci
  18(11),  1664--1671 (2015), article

\bibitem{Fisher36}
Fisher, R.A.: The use of multiple measurements in taxonomic problems. Annals of
  Eugenics  7(7),  179--188 (1936)

\bibitem{Fournier2016}
{Fournier}, N., {Delattre}, S.: {On the Kozachenko-Leonenko entropy estimator}.
  ArXiv e-prints  (Feb 2016)

\bibitem{Glasser2016}
Glasser, M.F., Coalson, T.S., Robinson, E.C., Hacker, C.D., Harwell, J.,
  Yacoub, E., Ugurbil, K., Andersson, J., Beckmann, C.F., Jenkinson, M., Smith,
  S.M., Van~Essen, D.C.: A multi-modal parcellation of human cerebral cortex.
  Nature  536(7615),  171--178 (2016), article

\bibitem{Goodfellow14}
Goodfellow, I., Pouget-Abadie, J., Mirza, M., Xu, B., Warde-Farley, D., Ozair,
  S., Courville, A., Bengio, Y.: Generative adversarial nets. In: Advances in
  Neural Information Processing Systems 27, pp. 2672--2680 (2014)

\bibitem{Gymrek2013}
Gymrek, M., McGuire, A.L., Golan, D., Halperin, E., Erlich, Y.: Identifying
  personal genomes by surname inference. Science  339(6117),  321--324 (2013)

\bibitem{Harmanci2016}
Harmanci, A., Gerstein, M.: Quantification of private information leakage from
  phenotype-genotype data: linking attacks. Nat Meth  13(3),  251--256 (2016)

\bibitem{Hayes2017}
{Hayes}, J., {Melis}, L., {Danezis}, G., {De Cristofaro}, E.: {LOGAN:
  Evaluating Privacy Leakage of Generative Models Using Generative Adversarial
  Networks}. ArXiv e-prints  (2017)

\bibitem{Huang17}
Huang, C., Kairouz, P., Chen, X., Sankar, L., Rajagopal, R.: Context-aware
  generative adversarial privacy. CoRR  abs/1710.09549 (2017)

\bibitem{Kairouz16}
Kairouz, P., Bonawitz, K., Ramage, D.: Discrete distribution estimation under
  local privacy. CoRR  abs/1602.07387 (2016)

\bibitem{Kos17}
Kos, J., Fischer, I., Song, D.: Adversarial examples for generative models.
  CoRR  abs/1702.06832 (2017)

\bibitem{Lee17}
Lee, H., Han, S., Lee, J.: Generative adversarial trainer: Defense to
  adversarial perturbations with {GAN}. CoRR  abs/1705.03387 (2017)

\bibitem{Liang17}
Liang, B., Li, H., Su, M., Li, X., Shi, W., Wang, X.: Detecting adversarial
  examples in deep networks with adaptive noise reduction. CoRR  abs/1705.08378
  (2017)

\bibitem{vanDerMaaten2008}
van~der Maaten, L., Hinton, G.: Visualizing data using {t-SNE}. Journal of
  Machine Learning Research  9,  2579--2605 (2008),
  \url{http://www.jmlr.org/papers/v9/vandermaaten08a.html}

\bibitem{Mirjalili2018}
Mirjalili, V., Raschka, S., Namboodiri, A.M., Ross, A.: Semi-adversarial
  networks: Convolutional autoencoders for imparting privacy to face images.
  CoRR  abs/1712.00321 (2017)

\bibitem{Mirjalili2017}
Mirjalili, V., Ross, A.: Soft biometric privacy: Retaining biometric utility of
  face images while perturbing gender. IJCB pp. 564--573 (2017)

\bibitem{Narayanan08}
Narayanan, A., Shmatikov, V.: Robust de-anonymization of large sparse datasets.
  In: Security and Privacy, 2008. SP 2008. IEEE Symposium on. pp. 111--125.
  IEEE (2008)

\bibitem{NIST}
NIST: Nist 8-bit gray scale images of fingerprint image groups (figs)

\bibitem{Oh2017}
Oh, S.J., Fritz, M., Schiele, B.: Adversarial image perturbation for privacy
  protection - {A} game theory perspective. CoRR  abs/1703.09471 (2017)

\bibitem{Orekondy_2018_CVPR}
Orekondy, T., Fritz, M., Schiele, B.: Connecting pixels to privacy and utility:
  Automatic redaction of private information in images. In: The IEEE Conference
  on Computer Vision and Pattern Recognition (CVPR) (June 2018)

\bibitem{Rajpurkar2017}
{Rajpurkar}, P., {Hannun}, A.Y., {Haghpanahi}, M., {Bourn}, C., {Ng}, A.Y.:
  {Cardiologist-Level Arrhythmia Detection with Convolutional Neural Networks}.
  ArXiv e-prints  (2017)

\bibitem{Raval2017}
Raval, N., Machanavajjhala, A., Cox, L.P.: Protecting visual secrets using
  adversarial nets. In: CVPR Workshop Proceedings (2017)

\bibitem{Sun2017}
Sun, Q., Ma, L., Oh, S.J., Gool, L.V., Schiele, B., Fritz, M.: Natural and
  effective obfuscation by head inpainting. CoRR  abs/1711.09001 (2017)

\bibitem{Sweeney2013}
Sweeney, L., Abu, A., Winn, J.: Identifying participants in the personal genome
  project by name {(A} re-identification experiment). CoRR  abs/1304.7605
  (2013)

\bibitem{Tripathy17}
Tripathy, A., Wang, Y., Ishwar, P.: Privacy-preserving adversarial networks.
  CoRR  abs/1712.07008 (2017)

\bibitem{Trzcinski2012}
Trzcinski, T., Lepetit, V.: Efficient discriminative projections for compact
  binary descriptors. In: European Conference on Computer Vision. pp. 228--242.
  Springer, Berlin, Heidelberg (2012)

\bibitem{Wang2004}
Wang, Z., Bovik, A.C., Sheikh, H.R., Simoncelli, E.P.: Image quality
  assessment: from error visibility to structural similarity. IEEE Transactions
  on Image Processing  13(4),  600--612 (April 2004)

\bibitem{DBLP:journals/corr/ZhaoGFK15}
Zhao, H., Gallo, O., Frosio, I., Kautz, J.: Loss functions for neural networks
  for image processing. CoRR  abs/1511.08861 (2015),
  \url{http://arxiv.org/abs/1511.08861}

\end{thebibliography}
\end{document}